\documentclass{article}
\usepackage{microtype}
\usepackage{graphicx}
\usepackage{subfigure}
\usepackage{booktabs} 
\usepackage{footnote}
\usepackage{caption}

\usepackage[hidelinks]{hyperref}
\usepackage{float}
\usepackage{algorithmic}
\usepackage{multirow}

\usepackage[final]{sods_2023}

\usepackage{amsmath}
\usepackage{amssymb}
\usepackage{mathtools}
\usepackage{amsthm}

\DeclareMathOperator*{\argmin}{arg\,min}

\graphicspath{./Plots/}

\begin{document}
\title{
    Minibatching Offers Improved Generalization Performance for Second Order Optimizers
    (ICML 2023)
}
\author{
Eric Silk\thanks{Equal First Authors} \\
    Department of Electrical \& Computer Engineering \\
    University of Illinois Urbana-Champaign, Urbana, IL, USA \And
Swarnita Chakraborty\footnotemark[1]\\
    Department of Mathematics \& Statistics \\
    Washington State University \\
    Pullman, WA, USA \AND
Nairanjana Dasgupta \\
    Department of Mathematics \& Statistics \\
    Washington State University \\
    Pullman, WA, USA \And
Anand D. Sarwate \\
    Department of Electrical \& Computer Engineering \\
    Rutgers, The State University of New Jersey \\
    New Brunswick, New Jersey, USA \And
Andrew Lumsdaine \\
    Paul G. Allen School of Computer Science \\
    University of Washington \\
    Seattle, WA, USA \And
Tony Chiang\thanks{Corresponding Author}\\
    Pacific Northwest National Laboratory \& \\
    Department of Mathematics \\
    University of Washington, Seattle, WA, USA
}

\maketitle

\begin{abstract}
Training deep neural networks (DNNs) used in modern machine learning is computationally expensive.  Machine learning scientists, therefore, rely on stochastic first-order methods for training, coupled with significant hand-tuning, to obtain good performance.  
To better understand performance variability of different stochastic algorithms, including
second-order methods, we conduct an empirical study that treats performance
as a response variable across multiple training sessions of the same model.
Using 2-factor Analysis of Variance (ANOVA) with interactions, we show that batch size
used during training
has a statistically significant effect on the peak accuracy of
the methods, and that full batch largely performed the worst.  
In addition, we found that second-order optimizers (SOOs) generally exhibited significantly
lower variance at specific batch sizes, suggesting they may require less
hyperparameter tuning, leading to a reduced overall time to solution for model training.
\end{abstract}

\section{Introduction}
Training a neural network involves optimizing the model's parameters to minimize a
given loss function, typically achieved through backpropagation, where gradients of
the loss function with respect to model parameters are used to update the parameters.
First-order optimization algorithms such as stochastic gradient
descent (SGD) have been the state-of-the-art approach. 
Recently, however, there has been growing interest in the use of second-order
optimization for neural network
training, and their advantages and disadvantages have been studied from
various perspectives~\cite{second_order_optimization,adahessian_2021,kaisa_2021,Quasi_Gauss_Newton_2020}.
Second-order
optimization algorithms are theoretically more effective as they consider the
curvature of the loss function and can achieve quadratic convergence in the vicinity
of a minimum.  
It is commonly believed, however, that they require more computational and memory resources, 
making them impractical for the larger problem sizes of contemporary models.
This view is supported both by  research~\cite{Goodfellow-et-al-2016} and 
practice~\cite{stackexchange-soo-a,stackexchange-soo-b}
(although there is some debate on this matter~\cite{stackexchange-soo-positive}).

%


In this paper, we use a two-way Analysis of Variance (ANOVA) model to
test the hypothesis that performance of three derivative-based
training algorithms (SGD plus two second-order methods) have the same relative
performance in terms of peak accuracy of the models, both with and without
minibatching.
This analysis provides an example of how we can use statistical techniques
for experimental design and analysis to understand the training of machine learning
models.
Our experiments tend to support the claim that randomization via minibatching can be
beneficial for both first and second-order optimization methods. This is somewhat
surprising because stochastic-based optimization would be expected to perform
relatively similarly to full batch optimization.
Our experiments gives empirical evidence to the contrary, showing that stochasticity
driven by minibatching is beneficial not only for gradient descent but also for the
two second-order methods we include. This suggests that neural network training
behaves differently than problems in classical optimization.

\section{Contributions}
\begin{itemize}
    \item We provide a rigorous statistical framework to compare the performance between optimizers in training deep neural networks. 
    \item We show that mini-batching not only has a regularizing effect for SGD but also for optimizers incorporating second order information.
    \item We provide empirical evidence for an optimal batch size across these optimizers. This evidence also shows that inclusion of second order information at this optimal batch size desensitizes model training to common hyperparameter tuning.
\end{itemize}

\section{Related Work}
\paragraph{Levenberg-Marquardt} Previous studies used methods such as
Levenberg-Marquardt~\cite{lecun_gradient_based_learning,lecun2012efficient},
but made modifications (typically diagonal approximation of the Hessian) to account
for perceived issues with the stochastic effects of minibatching. 
\paragraph{Limited-memory Broyden-
Fletcher-Goldfarb-Shann} In another study, limited-memory Broyden-Fletcher-Goldfarb-Shanno
(L-BFGS)~\cite{doi:10.1137/0916069} and conjugate gradient (though not nonlinear
conjugate gradient) were compared to SGD on two small models:
 an autoencoder with approximately $8\times 10^5$
parameters and a convolutional neural network with less than $10^4$ parameters~\cite{le2011optimization}.

\paragraph{Comparative Analysis} Results in the literature that contrast SGD and second-order methods
have generally been empirical studies comparing model accuracy and training time based on
specific instances of these optimizers using the MNIST dataset~\cite{deng2012mnist}. 
While prior studies found a preference for larger minibatch sizes (where SGD required extensive manual tuning while the other optimizers may prove to
be more robust),
studies typically report only the best or worst results of numerical experiments, often without
reproducible code nor a sufficiently detailed experimental design
to permit accurate assessments.  As a result, there has been little to no statistical
analysis to support the advantage of one approach over the other.

\section{Methodology}

Our experiments were carried out using Python 3 with the PyTorch machine learning
framework~\cite{pytorch}. The implementation is freely available on GitHub~\cite{silk_2022_anon}).  
All experiments were conducted on an Nvidia DGX-2 containing 16 Nvidia V100 GPU's,
utilizing one GPU per training session.

\subsection{Dataset, Model Architecture, and Training Regime}

We used the ResNet18 model~\cite{resnet} available in PyTorch. This model was selected as it
balanced the improved generalizability of our conclusions compared to simpler models (such as small
multi-layer perceptrons) against higher computational costs, affording more experiments to improve
the likelihood of a statistically significant result. A pre-trained model was not used, and the
number of outputs was set to 10 to match the number of classes in the dataset.

The CIFAR10 dataset~\cite{cifar10} was selected for the same reasons as model selection. A
training-test split of 50,000/10,000 was used. PyTorch's built-in data loader was used, which
selects batches of a fixed size randomly without replacement. Batch size was treated as a
hyperparameter, and values were selected intentionally such that they were integer divisors of the
full training set, guaranteeing there were no truncated batches (i.e., $n\!\mod k=0$ where $n$ is the
training set size and $k$ is the minibatch size). This applied to both gradient descent and the second-order optimizers. 
Training consisted of 100 epochs, each consisting of a train and test cycle, except in the case of
%
early termination due to an error condition or instability in the optimizer.
These results were \emph{not discarded} automatically, as they may have exhibited high accuracy
before the failure and therefore removing them could introduce unintended bias into the results. Experimental
specifications were generated by selecting
discrete sets of ``reasonable'' values for various hyperparameters (for instance, momentum could
take the value 0, 0.1, 0.5 and 0.9 for SGD) and taking the Cartesian product of all hyperparameter
sets for a given optimizer to generate a collection of hyperparameter combinations (experimental setups). On a per-optimizer basis, a setup was selected from this collection randomly without replacement and trained.  The intent is
to resemble a randomized grid search hyperparameter tuning process, such as might be conducted when
an end-user is testing a new model or optimizer without \textit{a priori} knowledge of
hyperparameter selection.

\subsection{Optimization Algorithms}

The gradient descent algorithm used was PyTorch's built-in implementation. When momentum was enabled (i.e., momentum was
$>0$), it was not Nesterov momentum. All other arguments were left with their default values.  
Dampening and weight decay were
set to 0, \texttt{maximize} was set to false, \texttt{foreach} was set to none, and \texttt{differentiable} was set to false.
\begin{figure*}
    \centering
    \includegraphics[width=\textwidth]{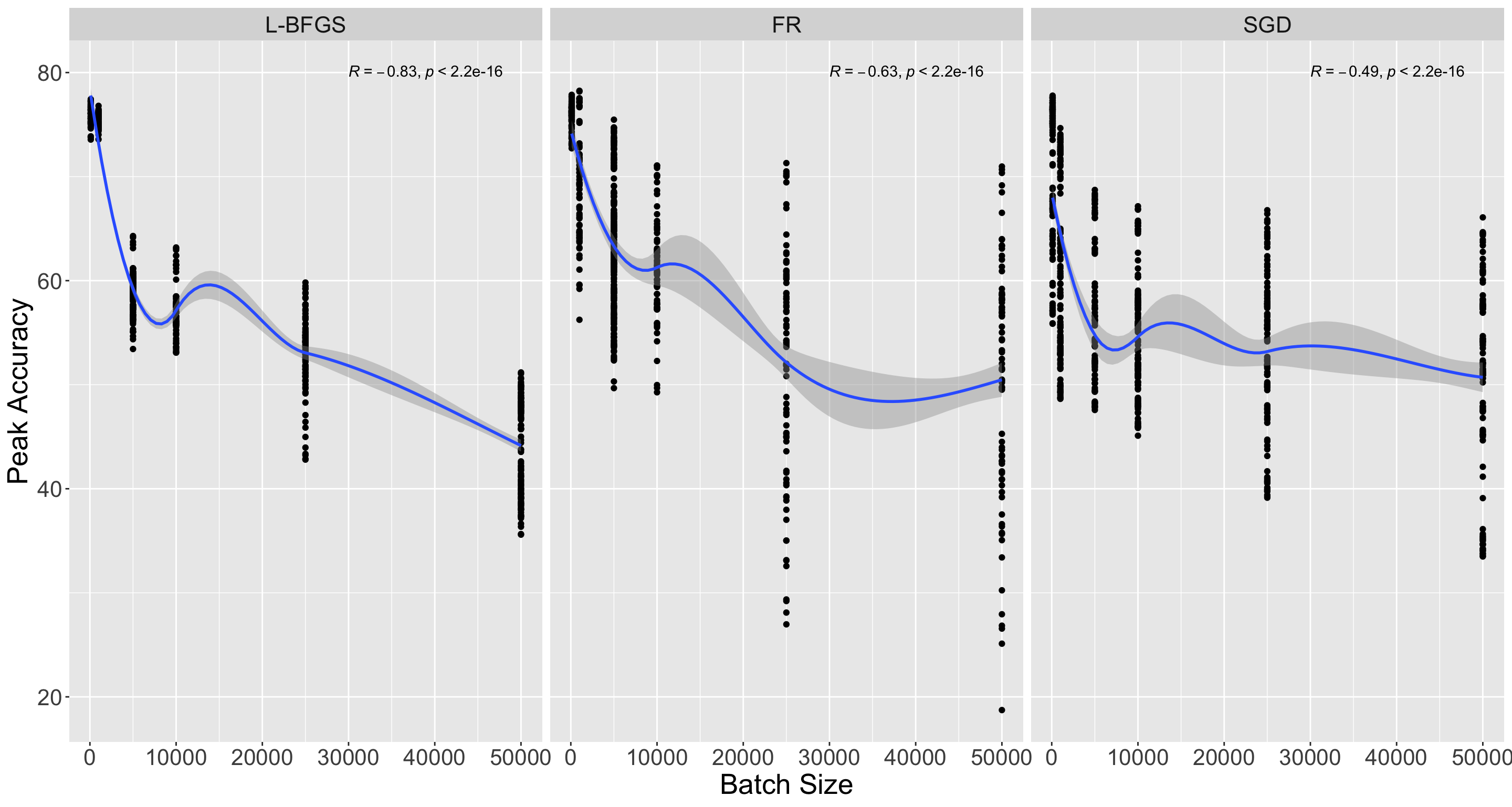}
    \caption{Peak accuracy trend for different batch Sizes for L-BFGS, FR and SGD optimization
        methods with outlier treatment}
    \label{fig:peak_acc_trend}
\end{figure*}

This effectively resulted in the following algorithm:

\begin{algorithmic}
\FOR{i=1\ldots n}
    \STATE $g_i\gets \nabla_{\theta} f_i(\theta_{i-1})$
    \IF{$\mu\neq 0$}
        \IF{$i>1$}
            \STATE $b_i\gets\mu b_{i-1} + (1-\tau)g_i$
        \ELSE
            \STATE $b_i\gets g_i$
        \ENDIF
        \STATE $g_i\gets b_i$
    \ENDIF
    \STATE $\theta_i\gets\theta_{i-1} - \gamma g_{i}$
\ENDFOR
\end{algorithmic}
Here, $\gamma$ is the learning rate, $\theta_0$ is the model parameters, $f(\theta)$ is the objective function,
and $\mu$ is the momentum ($0$ implies it is disabled).

Details for the second-order optimizers can be found in the appendix.

\subsection{Analysis of Variance}

The objective of this study was to determine if there is statistical evidence to demonstrate
any performance differences between second-order optimization and SGD (first-order optimization) 
for neural network model training, subject to different batch sizes.
We use peak accuracy as the performance metric, with
L-BFGS, Fletcher-Reeves (FR)~\cite{fletcher_reeves}, and SGD as the optimizers.
To understand the performance of each of the three optimization methods with respect
to batch size, we have conducted a two-way Analysis of Variance (ANOVA) model with
batch-size and the optimizers as the fixed treatment effects allowing for potential
interactions and peak-accuracy as the response.

In particular, we applied ANOVA to determine if there is a statistically detectable
interaction effect between optimizers and the batch size, i.e., if there is a
synergistic or antagonistic effect on the peak accuracy of the three optimizers for
different batch sizes.
Subsequent analysis depends upon the presence of detectable interactions.
\begin{itemize}
\item  If the two-way ANOVA model reports a detectable interaction effect, the marginal
  effects are not easily interpretable. As a next step, we conduct pair-wise tests for all
  combinations of batch sizes and optimizers to find out which pairs are significantly
  different in terms of peak accuracy
\item If there is no detectable interactions we could look at a marginal effect of batch size and optimizers and compare them within the groups.
 \end{itemize}
Since the distribution of peak accuracy for each of
the methods were right-skewed (the FR method showed slight bi-modality), we conducted our analysis based on the natural
logarithm of the peak accuracies. In general Log transformations are also the variance stabilizing transformations and this allowed us to deal with potentially unequal variance issues across the optimizers.  

Our experiments did include instances where the neural network model reported $10\% -15\%$ model
accuracy, suggesting that it might not be performing better than a random classifier for CIFAR10.
Because ANOVA results are affected by the presence of outlying values, we exclude these
in the main results for this paper (the majority of such cases occurred in the FR method at
batch size 5000 (see Figure \ref*{fig:box_plot})) though we do include these outliers in a parallel
analysis detailed in the Appendix.
For a detailed review of ANOVA and the model equations, we refer the reader to~\cite{anova_larson2008analysis}
and~\cite{anova_scheffe1999analysis}.

\begin{figure*}
    \centering
    \includegraphics[width=\textwidth]{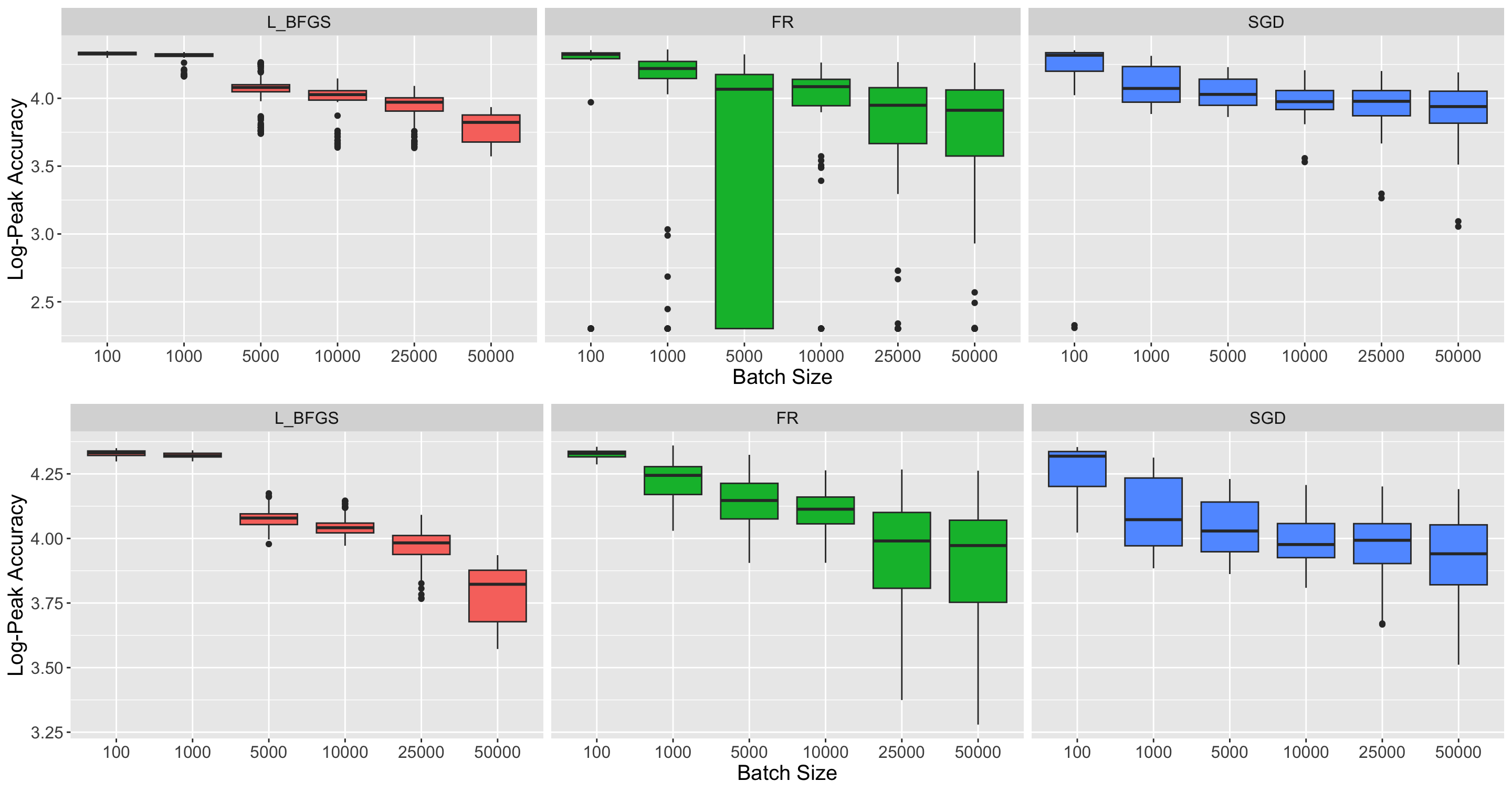}
    \caption{Box Plot of Log-Peak accuracy for different batch Sizes for L-BFGS, FR and SGD methods.
        Top: with outliers, bottom: treated for outliers}
    \label{fig:box_plot}
\end{figure*}

\section{Results}
Our overall results from ANOVA are given in Table~\ref*{table:anova_results_treated}, where we can see detectable interaction effects (p-value .0001).  To understand the interaction effects we looked at pairwise differences for the different treatments within each batch size.  Figure~\ref*{fig:box_plot} shows the box plots of the log peak accuracy for the treatments across the batch sizes and we see that that there are some variance difference across the treatment combinations with some combinations.  For L-BFGS and  FR at smaller batchsizes  showing much smaller variation than the larger batch sizes.

Our main results are summarized in the following sections.

\subsection{Minibatching Gradient-Based Optimizers Increases Test Accuracy as the Batch Size Decreases}
Our experiments demonstrated that peak accuracy for each of the three optimization methods
decreases with batch size. A correlation test reported p-values for each of three methods
to be strictly less than $0.05$; however a strict linear trend was not guaranteed
(see Figure~\ref*{fig:peak_acc_trend}).  While this is true overall, the interaction is explained by the difference in what happens at the different batch sizes across the methods. 

\subsection{Full-batch Second-order Optimizers Do Not Outperform Full-batch Gradient Descent}
Both full-batch L-BFGS and full-batch FR demonstrate lower peak test accuracy compared to full-batch
gradient descent (i.e., no stochasticity for back-propagation), and the full-batch SOOs' poorer performance becomes more exaggerated relative to stochastic gradient descent (Table \ref*{table:pairwise_treated}). Each SOO exhibits a slightly different
degradation in convergence results.  L-BFGS produces a lower mean peak performance even though it displays the tightest variability amongst the three optimizers; conversely, FR produces roughly the same mean when compared with gradient descent (close enough to be in the
same group when using pairwise testing for interaction effects), but contained the largest variability. Taken holistically, the claim that SOOs may perform better than first-order optimizers with an up-front computational penalty is not empirically supported.

\begin{table*}
    \caption{ANOVA Results of Log Peak Accuracy}
    \centering
    \begin{tabular}{ |p{1.8cm}|p{0.5cm}|p{1.8cm}|p{2.0cm}|p{1.8cm}|p{1.2cm}| }
        \hline
        \multicolumn{6}{|c|}{ANOVA Table }                               \\
        \hline
        Source      & DF & Type III SS & Mean Square & F Value & $Pr>F$   \\
        \hline
        Batch Size  & 5  & 35.92      & 7.18        & 480.66  & $<0.001$ \\
        Optimizer   & 2  & 1.37       & 0.684       & 45.78   & $<0.001$ \\
        Interaction & 10 & 4.11       & 0.41        & 27.51   & $<0.001$ \\
        \hline
    \end{tabular}
    
 \footnotesize{$^*$DF is the degree of freedom. Type III SS is Type 3 Sum of Squares. $`Pr > F'$ is the p-value.} 
    \label{table:anova_results_treated}
\end{table*}

\subsection{At Smaller Batchsizes SOOs  Outperform SGD}
When considering mean peak test accuracy for a fixed minibatch, second-order optimizers generally outperform
SGD at smaller batch sizes. Below a batch size of 25000, L-BFGS and FR consistently produce means that are either
statistically indistinguishable but slightly higher, or statistically significant improvements. This
is most apparent at a batch size of 1000: all three are significantly different than the others, and
SGD performs the worst of the three. At 25000, all three are statistically the same, but SGD
produces a slightly high mean than FR, while L-BFGS is still the highest. For a batch size of 100, there is no statistical differentiation between the mean peak accuracies of the three optimizers though 
both SSOs have variance that is considerably tighter than SGD (Table \ref*{table:pairwise_treated}). In fact, L-BFGS's variance is tighter even relative to that of FR due to the absence of any outliers for that batch size.  

\begin{table*}
    \caption{Pairwise Tests for Interaction Effects}
    \centering
    \begin{tabular}{ |p{2cm}|p{2cm}|p{2cm}|p{3cm}| }
        \hline
        \multicolumn{4}{|c|}{Pairwise Tests for Interaction Effects} \\
        \hline
        Batch Size & Treatment & Least Square Means & Letter Grouping$^*$      \\
        \hline
        100        & L-BFGS    & 4.3292   & A                        \\
        100        & FR        & 4.3256   & A                        \\
        100        & SGD       & 4.2597   & A                        \\
        \hline
        1000       & L-BFGS    & 4.3222   & A                        \\
        1000       & FR        & 4.2354   & B                        \\
        1000       & SGD       & 4.0879   & C                        \\
        \hline
        5000       & FR        & 4.1471   & A                        \\
        5000       & L-BFGS    & 4.0456   & B                        \\
        5000       & SGD       & 4.0358   & B                        \\
        \hline
        10000      & FR        & 4.2354   & A                        \\
        10000      & L-BFGS    & 4.0456   & A                        \\
        10000      & SGD       & 3.9925   & B                        \\
        \hline
        25000      & L-BFGS    & 3.9687   & A                        \\
        25000      & SGD       & 3.9644   & A                        \\
        25000      & FR        & 3.9288   & A                        \\
        \hline
        50000      & SGD       & 3.9089   & A                        \\
        50000      & FR        & 3.8894   & A                        \\
        50000      & L-BFGS    & 3.782    & B                        \\
        \hline
    \end{tabular}
    \\

    \footnotesize{$^*$ Different letters denote a statistical significance between groups }
    \label{table:pairwise_treated}
\end{table*}




\subsection{Minibatched Second-order Optimizers Trade Slower Convergence for Better ``Typical'' Convergence}
\begin{figure*}
    \centering
    \includegraphics[width=\textwidth]{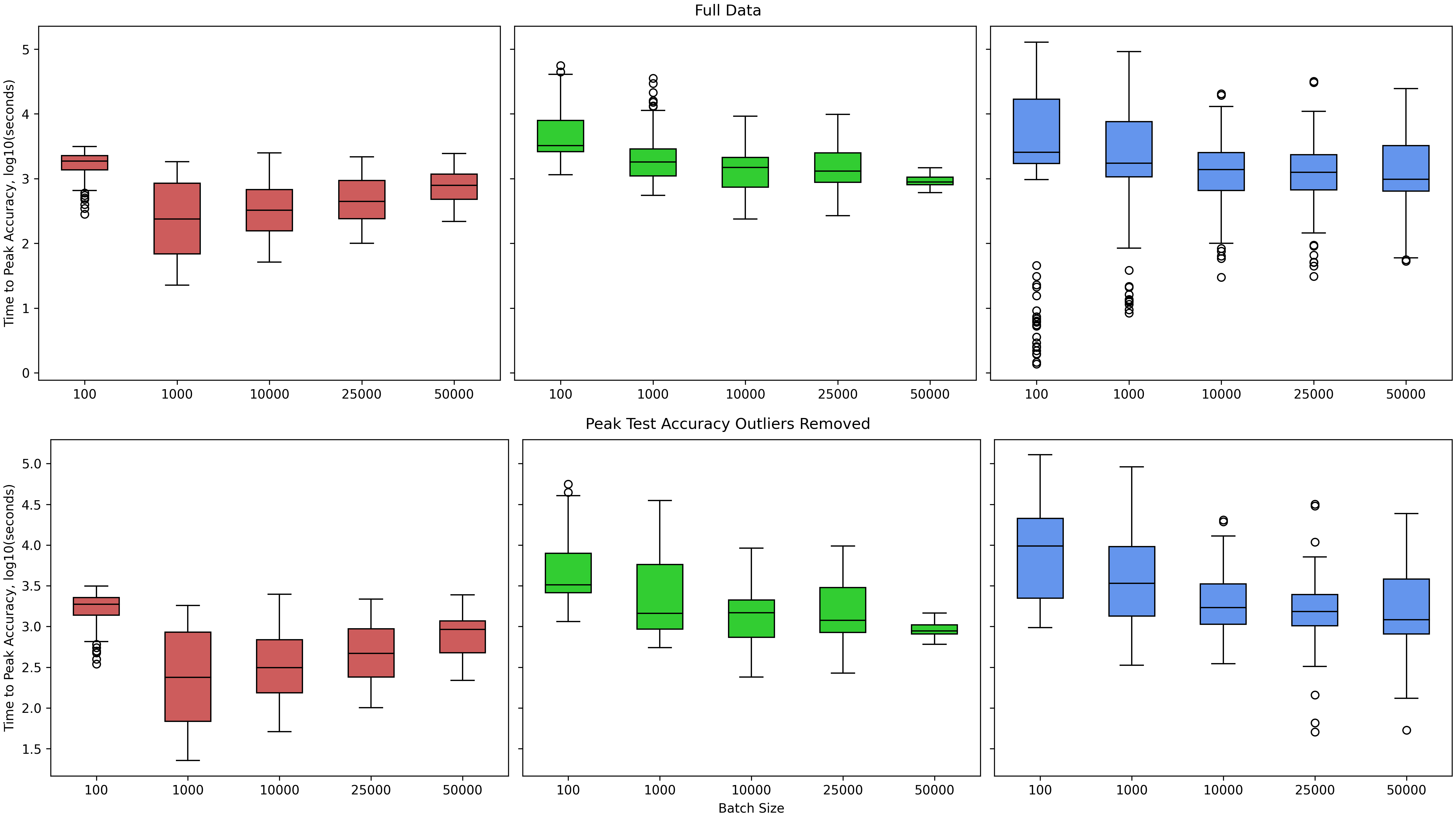}
    \caption{Ratio between Time to Peak Accuracy per optimizer per batch size. Top: Untreated data, Bottom: Peak Accuracy Outliers Removed}
    \label{fig:ttp_violin_comparison}
\end{figure*}

While minibatched SOOs with optimal batch size appear to win in terms of typical peak accuracy and decreased variance in peak accuracy, they do so at the expense of longer typical times to converge. However, they do so at a
better rate of exchange than anticipated -- for a batch size of 100, L-BFGS requires a mean time of only 5.1 times and a median time of a mere 1.7 times longer than SGD with the same batch (see Figure~\ref*{fig:ttp_violin_comparison} and Table~\ref*{table:ttp_ratios}).

Similarly, we also found that exceptionally fast training times for the second-order methods correlated with
especially poor peak accuracy. This implies the optimizers are likely finding  highly sub-optimal local extrema and
converging to them very quickly (which is not unexpected given the convergence guarantees of most optimizers in a non-convex
setting) or simply failing to converge as the $10\%$ peak accuracy is not better than random classification for CIFAR10. In a real world scenario, these situations could be rapidly identified by end users (or programmatically
via some heuristics), and the training session restarted with new parameter initialization. 
This correlation strengthens our decision to remove the extremely poor performing outliers.

\section{Discussion}

We have identified five takeaway messages based on our experimental evidence. In total, they suggest a more complicated story regarding the benefits and drawbacks of second-order optimization: the choice of batch size can make a significant impact because it can control some of the variability in stochastic optimization. Furthermore, it is tempting to consider optimization performance in isolation by comparing individual algorithms. However, if we consider entire workflows the considerations may become different because of variability across runs and hyperparameter tuning. Our work gives an example of how to understand these effects using an effective experimental design and statistical analysis can be beneficial.

\subsection{Acceleration via Larger Batch Sizes is Likely Not a viable Path Forward}
There was some hope that SOOs may show faster per-epoch convergence at larger batch sizes, which
would imply that, while they may require more calculations per iteration, hardware parallelization
could be utilized to provide time-to-solution
convergence improvements over the more serial process of
minibatching.  Unfortunately, given the aforementioned trend between batch size and convergence,
this does not appear feasible under current training regimes. As such, fundamental changes in
optimization algorithms or training regimes will need to be explored.

\subsection{Minibatching Offers Improved Convergence and Decreased Variance with Optimal Batch Size, Regardless of Optimizer Type}
One of the strongest trends exhibited in this study is that there appears to be an optimal batch size that will
improve accuracy and decrease the variance of the final test accuracy that is orders of magnitude lower than the
size of the full dataset. This effect appears to generalize beyond first-order gradient descent to second-order
methods without any attempt to account for stochasticity introduced.

Perhaps most interestingly, the second-order methods (and L-BFGS in particular) produce far lower
variance in their distributions. From the evidence, we can infer that, given an ``optimal'' batch size, second-order
optimizers may be insensitive to hyperparameter setting for fixed architecture and training data, and thus a generic training session is
highly likely to produce good convergence relative to the peak possible performance of the model.
This has several important ramifications, discussed in the following sections.

\subsection{Second-order Optimizers May Obviate Hyper-optimization, Producing Amortized Time to Solution Savings}

A common criticism of second-order methods is that they are too computationally expensive to be
worth using, and the results here support that when evaluating a single training session to another.
However, when evaluating the entire cost of training, including hyperparameter tuning
(or ``hyperoptimization''), time-to-solution savings may be realized. Recall the 100 sample batch
size: the mean time to peak accuracy (TTPA) was only about 5.1 times longer, and the median TTPA was a mere \textasciitilde 1.7
times greater (Table \ref*{table:ttp_ratios}). This implies that in the median case, having to run SGD twice
is more expensive than one run of L-BFGS; in the mean case it is 5-6 times more expensive.
    
\begin{table}
    \begin{tabular}{|l|l|l|llllll|}
        \hline
        \multicolumn{1}{|c|}{\multirow{2}{*}{\textbf{Measure}}} & \multicolumn{1}{c|}{\multirow{2}{*}{\textbf{Optimizer}}} & \multicolumn{1}{c|}{\multirow{2}{*}{\textbf{Outliers}}} & \multicolumn{6}{c|}{\textbf{Batch Size}}                                                                                                                                                                                                  \\ \cline{4-9}
        \multicolumn{1}{|c|}{}                                  & \multicolumn{1}{c|}{}                                    & \multicolumn{1}{c|}{}                                                     & \multicolumn{1}{c|}{\textit{100}}        & \multicolumn{1}{c|}{\textit{1,000}} & \multicolumn{1}{c|}{\textit{5,000}} & \multicolumn{1}{c|}{\textit{10,000}} & \multicolumn{1}{c|}{\textit{25,000}} & \multicolumn{1}{c|}{\textit{50,000}} \\ \hline
        \multirow{4}{*}{Min}                                    & \multirow{2}{*}{FR}                                      & Untreated                                                                 & \multicolumn{1}{l|}{0.005}               & \multicolumn{1}{l|}{0.370}          & \multicolumn{1}{l|}{0.152}          & \multicolumn{1}{l|}{0.586}           & \multicolumn{1}{l|}{0.307}           & 0.245                                \\ \cline{3-9}
                                                                &                                                          & Treated                                                                   & \multicolumn{1}{l|}{2.804}               & \multicolumn{1}{l|}{14.815}         & \multicolumn{1}{l|}{3.865}          & \multicolumn{1}{l|}{6.832}           & \multicolumn{1}{l|}{0.502}           & 0.245                                \\ \cline{2-9}
                                                                & \multirow{2}{*}{L-BFGS}                                    & Untreated                                                                 & \multicolumn{1}{l|}{4.064}               & \multicolumn{1}{l|}{24.343}         & \multicolumn{1}{l|}{5.720}          & \multicolumn{1}{l|}{4.642}           & \multicolumn{1}{l|}{2.652}           & 2.772                                \\ \cline{3-9}
                                                                &                                                          & Treated                                                                   & \multicolumn{1}{l|}{3.325}               & \multicolumn{1}{l|}{24.343}         & \multicolumn{1}{l|}{5.720}          & \multicolumn{1}{l|}{4.672}           & \multicolumn{1}{l|}{2.652}           & 2.772                                \\ \hline
        \multirow{4}{*}{Mean}                                   & \multirow{2}{*}{FR}                                      & Untreated                                                                 & \multicolumn{1}{l|}{9.892}               & \multicolumn{1}{l|}{12.600}         & \multicolumn{1}{l|}{4.211}          & \multicolumn{1}{l|}{5.924}           & \multicolumn{1}{l|}{3.960}           & 2.848                                \\ \cline{3-9}
                                                                &                                                          & Treated                                                                   & \multicolumn{1}{l|}{12.508}              & \multicolumn{1}{l|}{15.024}         & \multicolumn{1}{l|}{6.467}          & \multicolumn{1}{l|}{7.359}           & \multicolumn{1}{l|}{4.491}           & 3.315                                \\ \cline{2-9}
                                                                & \multirow{2}{*}{L-BFGS}                                    & Untreated                                                                 & \multicolumn{1}{l|}{5.154}               & \multicolumn{1}{l|}{8.974}          & \multicolumn{1}{l|}{6.149}          & \multicolumn{1}{l|}{3.903}           & \multicolumn{1}{l|}{3.559}           & 1.075                                \\ \cline{3-9}
                                                                &                                                          & Treated                                                                   & \multicolumn{1}{l|}{5.119}               & \multicolumn{1}{l|}{10.082}         & \multicolumn{1}{l|}{5.967}          & \multicolumn{1}{l|}{4.105}           & \multicolumn{1}{l|}{3.684}           & 1.066                                \\ \hline
        \multirow{4}{*}{Median}                                 & \multirow{2}{*}{FR}                                      & Untreated                                                                 & \multicolumn{1}{l|}{1.367}               & \multicolumn{1}{l|}{7.285}          & \multicolumn{1}{l|}{3.309}          & \multicolumn{1}{l|}{4.248}           & \multicolumn{1}{l|}{2.811}           & 1.237                                \\ \cline{3-9}
                                                                &                                                          & Treated                                                                   & \multicolumn{1}{l|}{5.186}               & \multicolumn{1}{l|}{14.296}         & \multicolumn{1}{l|}{5.273}          & \multicolumn{1}{l|}{5.448}           & \multicolumn{1}{l|}{3.288}           & 1.310                                \\ \cline{2-9}
                                                                & \multirow{2}{*}{L-BFGS}                                    & Untreated                                                                 & \multicolumn{1}{l|}{1.741}               & \multicolumn{1}{l|}{7.628}          & \multicolumn{1}{l|}{4.807}          & \multicolumn{1}{l|}{4.557}           & \multicolumn{1}{l|}{2.939}           & 1.121                                \\ \cline{3-9}
                                                                &                                                          & Treated                                                                   & \multicolumn{1}{l|}{1.723}               & \multicolumn{1}{l|}{6.089}          & \multicolumn{1}{l|}{4.600}          & \multicolumn{1}{l|}{4.701}           & \multicolumn{1}{l|}{2.563}           & 0.955                                \\ \hline
        \multirow{4}{*}{Max}                                    & \multirow{2}{*}{FR}                                      & Untreated                                                                 & \multicolumn{1}{l|}{40.806}              & \multicolumn{1}{l|}{50.224}         & \multicolumn{1}{l|}{19.377}         & \multicolumn{1}{l|}{8.224}           & \multicolumn{1}{l|}{14.652}          & 9.991                                \\ \cline{3-9}
                                                                &                                                          & Treated                                                                   & \multicolumn{1}{l|}{40.806}              & \multicolumn{1}{l|}{50.224}         & \multicolumn{1}{l|}{19.377}         & \multicolumn{1}{l|}{8.224}           & \multicolumn{1}{l|}{14.652}          & 9.991                                \\ \cline{2-9}
                                                                & \multirow{2}{*}{L-BFGS}                                    & Untreated                                                                 & \multicolumn{1}{l|}{17.799}              & \multicolumn{1}{l|}{19.365}         & \multicolumn{1}{l|}{11.998}         & \multicolumn{1}{l|}{3.701}           & \multicolumn{1}{l|}{4.510}           & 0.597                                \\ \cline{3-9}
                                                                &                                                          & Treated                                                                   & \multicolumn{1}{l|}{17.799}              & \multicolumn{1}{l|}{19.365}         & \multicolumn{1}{l|}{11.998}         & \multicolumn{1}{l|}{3.701}           & \multicolumn{1}{l|}{4.510}           & 0.597                                \\ \hline
    \end{tabular}
    \caption{Ratio of Time to Peak Accuracy of FR and L-BFGS vs.~SGD, showing Min/Max, Mean and Median across all batch sizes (lower is better)}
    \label{table:ttp_ratios}
\end{table}

Thus, it may be desirable to ``buy once, cry once''
with training: accept a slightly higher upfront temporal cost to provide greater guarantees
about model convergence, obviating 
the need for expensive, time-consuming, and potentially complex
hyperparameter optimization. This may be especially attractive for users with constrained computing
environments who may not be able to perform grid-search algorithms in a parallel fashion (e.g., small
companies and universities, independent researchers, or hobbyists without access to multi-GPU systems).
This both helps simplify the process and lower the barrier to entry. Further study may be able to
identify hyperparameter sub-spaces that produce good convergence while minimizing TTPA (i.e., which
hyperparameters minimize TTPA for the batch size of 100 and 1000?), much in the same way that it is
standard practice to constrain the learning rate of SGD to small (i.e., $<<1$) values.

\subsection{SOOs offer increased confidence in performance given optimal batch size}
ML scientists should be much more confident in the peak performance of a generic model or training regime
after completing a training session for an optimal batch size (in this study, batch $= 100$). Given the
relatively larger variance exhibited by SGD at batch-size 100, we cannot gain immediate confidence in
the results of a given training session.
Conversely, it is clear whether  good initial result indicates the potential 
for significant performance improvement with further training, or if the result is already optimal. Moreover, the generalizability of
these results to different models or different datasets is unclear.
Many papers  report only one peak accuracy for a given problem statement, without exploring (or not presenting) results for multiple
hyperparameters. An optimizer with wide variance of accuracies may not work well outside of a narrow
hyperparameter space.

Further studies may be able to identify robust regions of the
hyperparameter space that generalize across problem sets, further improving the utility of these optimizers.
As an example, note the apparently bimodal distribution of FR. Clearly, certain configurations resulted
in virtually no learning; conversely, other configurations produced good convergence. If these can be
predicted with minimal effort \textit{a priori}, they can be avoided.

\subsection{Empirical results require rigorous experimental design and analysis}
Finally, from a certain point of view, we can think of training these models as conducting experiments, and so when exploring the real-world properties of these algorithms, we need robust experimental design and
sound statistical analysis. Our work shows the limitations of characterizing one method as being superior to another.
The reality is more nuanced, and it is far more productive to discuss the whole distribution of model behaviors. 
Furthermore, na\"{\i}ve statistical analyses may lead to 
to different (and potentially incorrect) conclusions.  The empirical findings from a fairly straightforward experimental design has shown us the importance of batch size as a hyperparameter in deep learning.  Indeed, minibatching is a critical parameter that can be optimized for training,
thereby extending the notion that \textit{SGD implicitly regularizes}\cite{smith2021origin}\cite{barrett2022implicit}.

As we continue to investigate the structures that strongly influence training via designed simulation experiments, we can propose better modeling techniques in general. 
%
Unfortunately, such simulation experiments seem rare, and we hope our work will open up avenues for this type of research.  

\section{Conclusion}
The results presented in this paper support the widely held belief that SGD is an implicit regularizer.  
We have, however, also added to this area of research by demonstrating that not all minibatches are created equal. Our work firmly dispels the notion that full batch optimization, even when coupled with second-order information, can compete with minibatched versions. Specifically, we have observed that SGD with smaller batch sizes does not converge to the same solution as full gradient descent. This suggests that revisiting intuitions based on classical optimization theory is needed. In particular, 
some of the assumptions we make about the ambient space for classical optimization are not applicable to the loss landscape in deep learning.  

Our most surprising result shows that an optimal batch size seems to desensitize a fixed model from other hyperparameter settings, as evidenced by the reduction of variance in peak accuracy (since hyperparameters in our experiment were set by a random sweep). While we have discussed the potential of leveraging time-to-solution savings from this reduction in variance as an immediate consequence,  we argue that a more general theory is needed for mathematical guarantees. For instance, does the addition of second-order information to the optimizer imply that we need only tune the batch size ignoring the other hyperparameters? If so, this trade-off could have significant implications for training massive neural networks such as so-called ``foundation models.'' 
There are mathematical ramifications as well: the choice of hyperparameters makes the loss landscape a moduli space defining global geometric structures. In some sense, the randomness from different minibatch sizes 
affects the local geometric structure during each batch relative to the model's current location. It may be that certain alterations of the local geometry permits the optimizers to avoid ``slot canyons'' in preference for wide ``flat basins'' independent of the global structure. This might be a reason why using relatively small batch size in \textit{Adam} (which leverages first and second moments)  consistently outperforms other optimizers~\cite{kingma2017adam}.    

There are two natural and complementary directions to build on the work we present here: (1)~using our results as hypothesis generation for theory and (2)~refactoring our training framework to view batch size as a hyperparameter that has direct consequence for model performance rather than simply for model training speed-up. Because we only selected two SOOs (each from a distinct family of optimizers), we also realize that more simulation studies are needed to understand other families such as the Non-linear Conjugate Gradient and the Quasi-Newton methods. Our initial work shows that there is understanding to be gained from these experiments. 
%


\bibliographystyle{icml2023}
\bibliography{references}

\clearpage
\newpage
\appendix
\section*{Appendix}

\subsection*{Optimization Algorithms}
Two popular algorithms were studied and contrasted against (stochastic) gradient descent, namely:
\begin{enumerate}
    \item The Fletcher-Reeves Nonlinear Conjugate Gradient method (FR)\cite{fletcher_reeves}
    \item The Limited Memory Broyden-Fletcher-Goldfarb-Shanno method (L-BFGS)\cite{lbfgs}
\end{enumerate}
These were selected owing to their robust theoretical underpinnings and success in more traditional
optimization problems.

\subsubsection*{Fletcher-Reeves}
FR is a specific method within a family of methods known as ``nonlinear conjugate gradient''
methods; which proceed at each iteration as follows:
\begin{enumerate}
    \item Calculate the direction of steepest descent (i.e the gradient) $\Delta x_i$
    \item Calculate a scalar value $\beta_i$ (detailed below)
    \item Update the conjugate direction $s_i=\Delta x_i+\beta_is_{n-1}$
    \item Perform a line-search $\alpha_i^* = \argmin_{\alpha} f(x_i+\alpha s_i)$
    \item Update $x_{i+1} = x_{i}+\alpha_is_i$
\end{enumerate}

Fletcher-Reeves calculates $\beta$ as:
\[\beta_i^{FR} = \frac{\|\Delta x_i\|_2}{\|\Delta x_{i-1}\|_2}\]
where $\|\cdot\|_2$ is the usual L2 vector norm. We additionally perform an automatic reset of
$\beta$ using the formula:
\[\beta\coloneqq\max(0, \beta)\]

A backtracking line-search was used for step 4, and the ``learning rate'' hyperparameter was used as
the initial guess for $\alpha$ instead of its more typical role as a damping coefficient, and any
value $\alpha>0$ was allowed (including $\alpha>1$).  Contraction rate and maximum number of line
searches were also made available as hyperparameters. In the event the line search failed (i.e.,
sufficient decrease was not found in the maximum number of searches) the last, and thus smallest,
value for $\alpha$ was used.

The number of steps taken per-batch is set as a hyperparameter and must be $>1$. Note that this
means no notion of conjugacy is retained between batches.

\subsubsection*{Limited Memory Broyden-Fletcher-Goldfarb-Shanno}
L-BFGS is a quasi-Newton method; that is, it relies on an approximation of the Hessian $B_i$ of the
system, rather than the true hessian. The algorithm proceeds as follows:
\begin{enumerate}
    \item Determine a step direction $p_i$ by solving $B_ip_i=-\Delta f(x_i)$
    \item Perform a line-search $\alpha_i^* = \argmin_{\alpha}f(x_i+\alpha p_i)$
    \item Set the temporary variable $s_i=\alpha_ip_i$
    \item Step s.t. $x_{i+1} = x_i+s_i$
    \item Take the difference of gradients $y_i=\Delta f(x_{i+1})-\Delta f(x_i)$
    \item Update $B_{i+1} = B_{i}+\frac{y_iy_i^T}{y_i^Ts_i}-\frac{B_is_is_i^TB_i^T}{s_i^TB_is_i}$
\end{enumerate}
where $B_0=I$. We  note that the algorithm does not need to actually materialize the matrix $B_i$, but rather it requires only the result of its product with a vector.  Consequently, the matrix-vector product can be represented as a recursive sum of intermediate matrix-vector products, as follows:
\[
    B_i*x
    = B_{i-1}x
    + \frac{y_{i-1}y_{i-1}^T}{y_{i-1}^Ts_{i-1}}x
    -\frac{B_{i-1}s_{i-1}s_{i-1}^TB_{i-1}^T}{s_{i-1}^TB_{i-1}s_{i-1}}x
\]
The matrix $B_0$ is typically taken to be the identity, i.e., the initial product is
$B_0x=x$. To realize this computation it is only necessary to 
retain the vectors $y_i$ and $s_i$. 
By choosing to only store the $n$ most recent vectors,
this method becomes the ``limited memory'' BFGS, or ``L-BFGS'' method.

The Wolfe line search found in SciPy's L-BFGS implementation was used~\cite{scipy}.
Similar to FR, learning rate is used as a maximum value for $\alpha$ and it can exceed 1.
Unlike FR, if the algorithm fails to find a suitable step size in the maximum number of
searches, no step is taken.

It is worth noting that, while there have been efforts to develop algorithms based around second-order methods  that account for or take advantage of the stochasticity introduced by minibatching~\cite{byrd_stochastic_lbfgs}, we do not make any such attempts in our implementation.

\newpage
\onecolumn
\section*{Results based on ANOVA on complete dataset}
\begin{table}[H]
    \caption{}
    \centering
    \begin{tabular}{ |p{1.8cm}|p{0.6cm}|p{1.8cm}|p{2.0cm}|p{1.6cm}|p{1.2cm}| }
        \hline
        \multicolumn{6}{|c|}{ANOVA with Complete Data}                   \\
        \hline
        Source      & DF & TypeIII SS & Mean Square & F Value & $Pr>F$   \\
        \hline
        BatchSize   & 5  & 67.82      & 13.56       & 27.01   & $<0.001$ \\
        Optimizer   & 2  & 75.31      & 37.65       & 74.99   & $<0.001$ \\
        Interaction & 10 & 25.65      & 2.57        & 5.11    & $<0.001$ \\
        \hline
    \end{tabular}
    
      \footnotesize{$^*$DF is the degree of freedom. Type III SS is Type 3 Sum of Squares. $`Pr > F'$ is the p-value.}
      \label{table:anova_with_outliers}
\end{table}

\begin{table}[H]
    \caption{Pairwise Tests for Interaction Effects}
    \centering
    \begin{tabular}{ |p{1.5cm}|p{2cm}|p{2cm}|p{3cm}| }
        \hline
        \multicolumn{4}{|c|}{Pairwise Tests for Interaction Effects} \\
        \hline
        Batch Size & Treatment & LS Means & Letter Grouping          \\
        \hline
        100        & L-BFGS    & 6.2458   & A                        \\
        100        & SGD       & 6.1036   & A                        \\
        100        & FR        & 5.6414   & B                        \\
        \hline
        1000       & L-BFGS    & 6.1904   & A                        \\
        1000       & SGD       & 6.1036   & AB                       \\
        1000       & FR        & 5.6794   & B                        \\
        \hline
        5000       & L-BFGS    & 5.866    & A                        \\
        5000       & SGD       & 5.8224   & A                        \\
        5000       & FR        & 5.1007   & B                        \\
        \hline
        10000      & SGD       & 5.75     & A                        \\
        10000      & L-BFGS    & 5.746    & A                        \\
        10000      & FR        & 5.4845   & A                        \\
        \hline
        25000      & SGD       & 5.704    & A                        \\
        25000      & L-BFGS    & 5.6706   & A                        \\
        25000      & FR        & 5.3751   & A                        \\
        \hline
        50000      & SGD       & 5.6183   & A                        \\
        50000      & L-BFGS    & 5.4562   & AB                       \\
        50000      & FR        & 5.253    & B                        \\
        \hline
    \end{tabular}
    
     \footnotesize{$^*$ Different letters denote a statistical significance between groups }
    \label{tab:pairwise_complete}
\end{table}

\begin{figure*}[H]
    \centering
    \includegraphics[width=\textwidth]{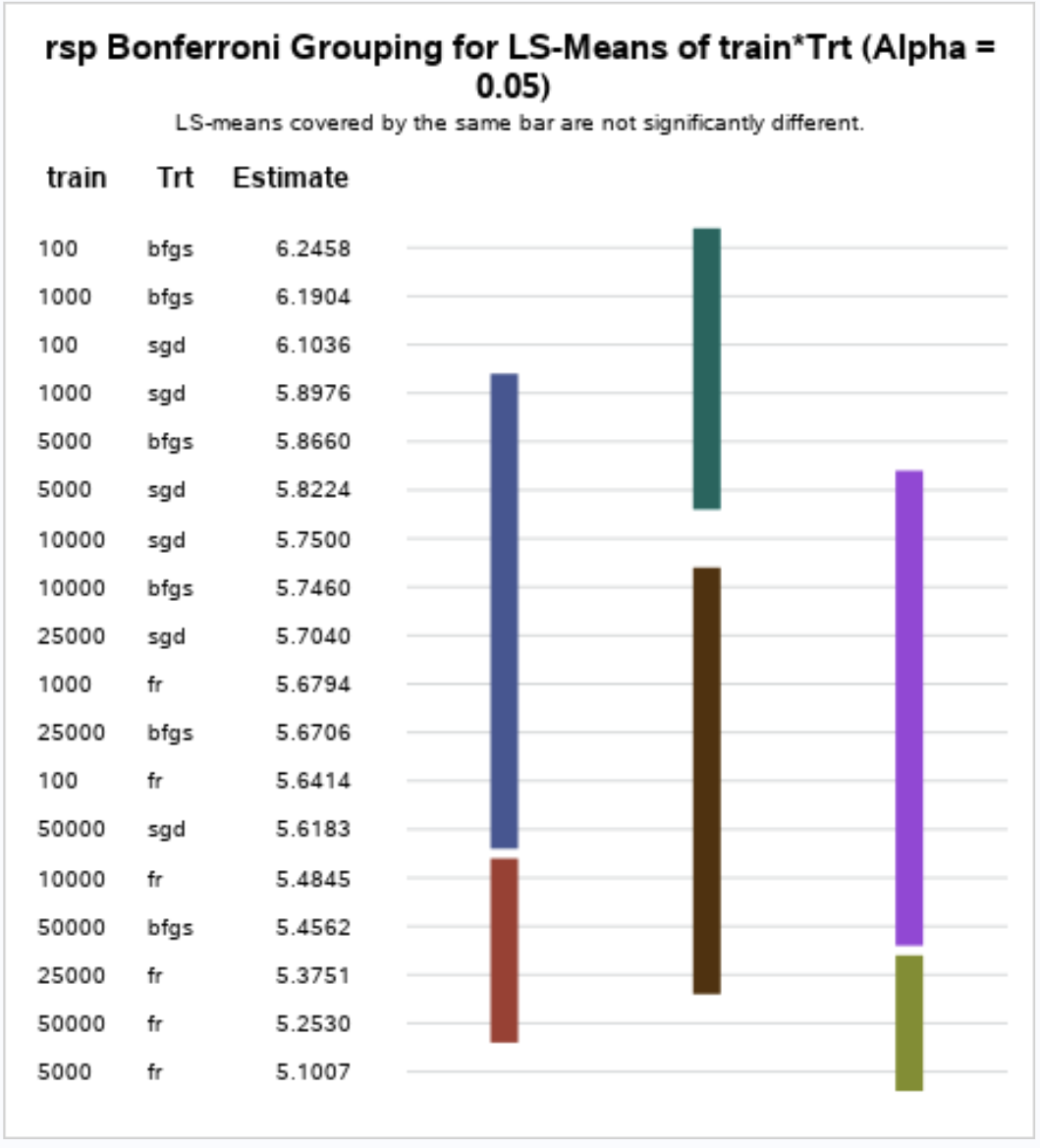}
    \caption{Pairwise Tests for the Interaction effect of Batch Size and Optimization Method based on complete data}
    \label{fig:pairwise_complete}
\end{figure*}

\newpage
\section*{Results with a Batch Size of 10}
\begin{figure*}[h]
    \centering
    \includegraphics[width=0.70\textwidth]{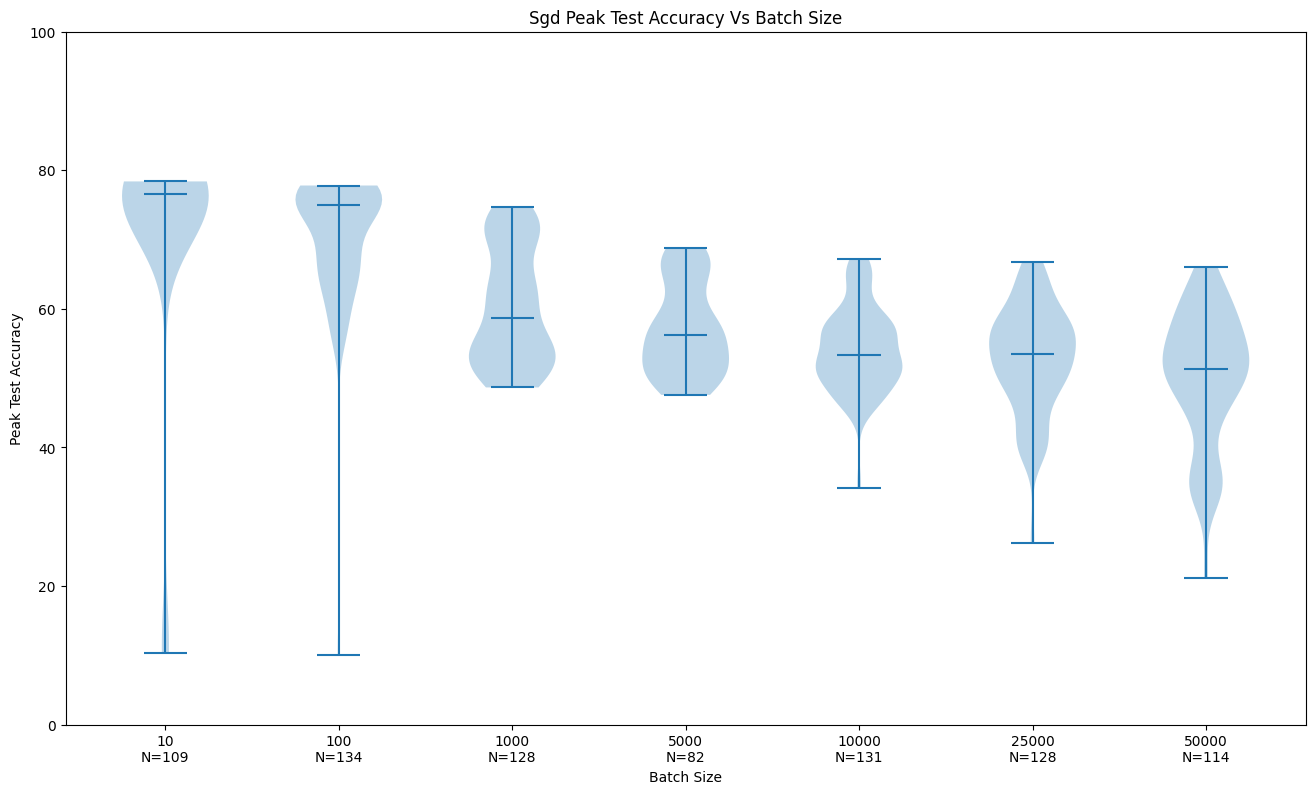}
    \caption{Full results of SGD, including batch size of 10}
    \label{fig:sgd_batch_10}
\end{figure*}

\begin{figure*}[h]
    \centering
    \includegraphics[width=0.70\textwidth]{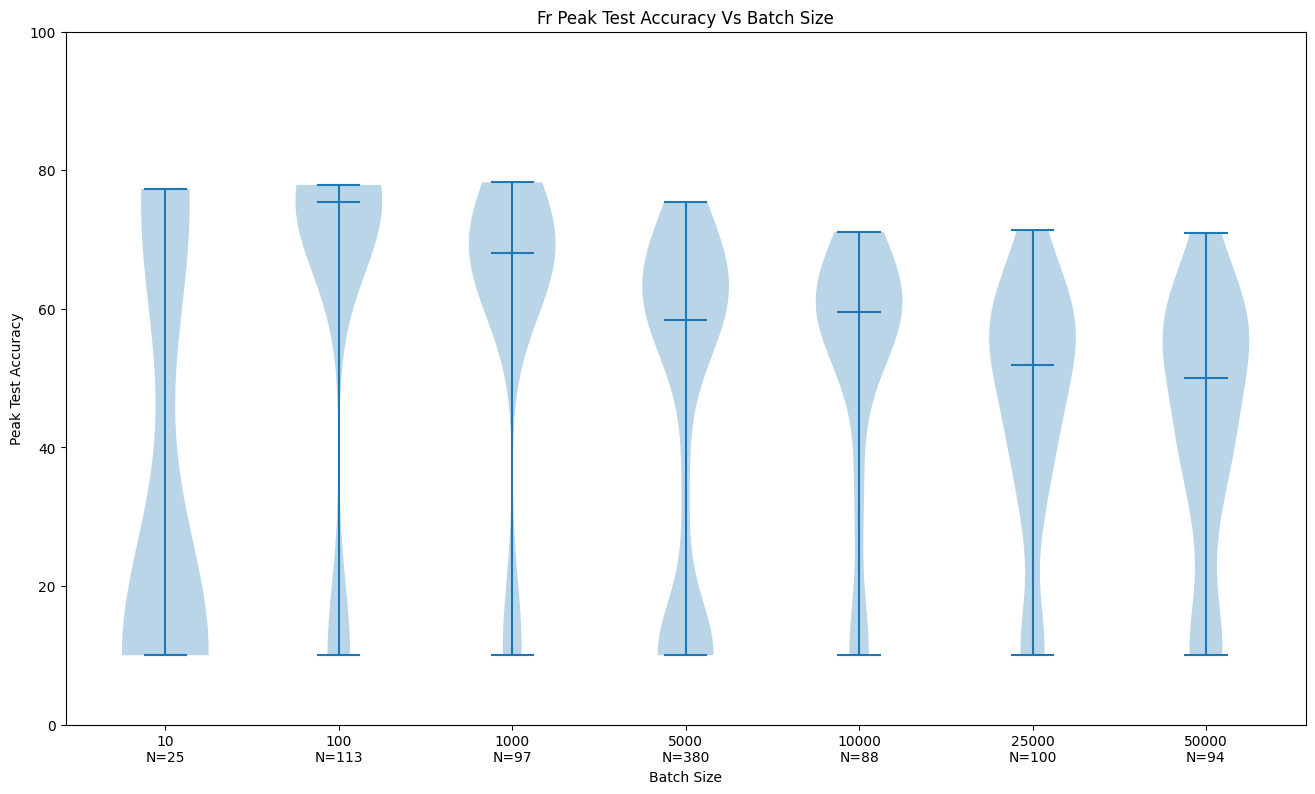}
    \caption{Full results of FR, including batch size of 10}
    \label{fig:fr_batch_10}
\end{figure*}

Several trials were conducted using a batch size of 10.  However, due to the excessive runtime required for batch sizes of 10, we opted not to conduct a complete
survey this space to provide more experimental data. Only SGD and FR were tested with this batch size, and their respective peak accuracy
violin plots are shown in Figures~\ref*{fig:sgd_batch_10} and \ref*{fig:fr_batch_10}.
Specifically, 109 runs were performed using SGD,  25 were completed for FR, and 0 for L-BFGS.
While rigorous analysis has not been conducted, note that the distribution of FR is markedly worse at this particular batch size, with a  ``bell-bottom'' shape rather than the expected 
wine glass shape.



\end{document}